# A Knowledge-Driven LLM-Based Decision-Support System for Explainable Defect Analysis and Mitigation Guidance in Laser Powder Bed Fusion

Basit Mahmud Shahriar[a], Md Habibor Rahman[a, 1]

[a] Department of Mechanical Engineering, University of Massachusetts Dartmouth, Dartmouth, MA 02747, USA

## Abstract

Identification and diagnosis of defects in metal additive manufacturing processes, such as Laser Powder Bed Fusion (LPBF), are knowledge-intensive tasks that require in-depth domain expertise encompassing defect ontologies, their causal relationships, and defect-specific control strategies. In practice, this knowledge is often scattered across the literature and difficult to integrate into decision-support workflows. While researchers have introduced Large Language Models (LLMs) and ontology-driven frameworks for information retrieval and decision support, existing approaches often rely on unstructured web-scale knowledge or static document retrieval and therefore provide limited support for domain-specific defect–process relationships, actionable mitigation guidance, and interactive defect exploration in manufacturing environments. To address these limitations, this work presents a knowledge-driven decision-support system that integrates structured defect knowledge with LLM-based reasoning to provide explainable defect diagnosis and mitigation guidance in manufacturing, using LPBF as a representative, safety-critical case study. The proposed ontology-integrated LLM-based decision support system for LPBF defect analysis and mitigation guidance is built on a knowledge base containing 27 known LPBF defect types organized into hierarchical categories and causal relationships. The developed system supports fuzzy natural language queries for systematic knowledge retrieval, literature-supported explanation of defects, and guidance on defect causes and mitigation strategies derived from encoded process knowledge. Furthermore, a multimodal image-assessment module based on foundation models enables descriptor-guided interpretation of representative microscopic defect images through semantic alignment scoring. The proposed framework was evaluated through qualitative comparisons with general-purpose vision-language models, an ablation study, and an inter-rater reliability analysis. Evaluation on the literature-derived dataset showed that the fully integrated configuration outperformed the other three evaluated system configurations, achieving a macro-average F1 score of 80.8%. Additionally, inter-rater reliability analysis using Cohen's kappa indicated substantial agreement ($k = 0.66$) between the model outputs and the literature-derived reference labels. These findings suggest that ontology-guided knowledge representation can improve the consistency, interpretability, and practical usefulness of LLM-assisted LPBF defect analysis. The proposed approach helps bridge the gap between fragmented research knowledge and practical shop-floor decision-making by providing a structured, explainable, and human-centered workflow for defect exploration, assessment, and guidance on defect mitigation.

## 1. Introduction

Laser powder bed fusion (LPBF) is one of the most widely adopted metal additive manufacturing (AM) processes for producing geometrically complex components for aerospace, biomedical, energy, and other high-performance applications. Despite its advantages, LPBF remains highly sensitive to process-induced defects, such as porosity, cracking, balling, surface roughness, layer delamination, and dimensional distortion [1]. These defects can compromise part integrity, repeatability, and qualification, and create significant challenges for process planning and quality assurance. In practice, effective diagnosis and mitigation of LPBF defects remain highly knowledge-intensive because relevant information is distributed across defect taxonomies, process studies, microscopy-based observations, and material-specific parameter investigations reported in the literature [2,3]. As a result, translating fragmented research knowledge into timely, actionable engineering guidance remains difficult, particularly for users and practitioners without deep domain expertise.

---

[1] Corresponding Author.
*E-mail address:* mrahman15@umassd.edu (Md Habibor Rahman).

In response, prior research has explored several relevant directions, including structured knowledge representations for defect modeling, data-driven and vision-based approaches for defect analysis, and more recent large language model (LLM)-based systems for technical querying and decision support. Recent studies have investigated structured knowledge representations, such as ontologies and knowledge graphs, to systematically represent LPBF defects and their associated causal knowledge, thereby enabling intelligent and explainable reasoning in LPBF defect analysis. Knowledge graph embedding techniques, for example, enable the representation of defects, process parameters, and causal relationships in machine-readable form, supporting classification and semantic querying [4]. These approaches are valuable because they impose conceptual structure on a domain characterized by inconsistent terminology, overlapping defect descriptions, and multilevel cause–and–effect relationships. However, knowledge-based representation approaches often rely on static query mechanisms and lack the interactive, flexible reasoning capabilities needed for real-time, human-centric decision support on the shop floor. In parallel, data-driven and vision-based approaches have been developed for defect detection, process optimization, and in situ monitoring [5,6]. Although these methods can achieve strong predictive performance, their black-box nature offers limited transparency and interpretability for defect diagnosis, root cause analysis, and mitigation guidance [7]. More recently, LLMs and retrieval-based systems have emerged as promising tools for technical querying, multimodal reasoning, and decision support in scientific and engineering contexts, as modern manufacturing paradigms increasingly emphasize human-centric AI systems that augment, rather than replace, expert decision-making [8–12]. LLM–based agents have been investigated for defect classification in microscopy images [9], decision support in healthcare [13], and the generation of natural language–driven simulation models [14], illustrating the growing capabilities of such agents to support reasoning and decision-making. While these developments improve accessibility and user interaction, their effective use in manufacturing remains constrained by the need for domain-specific knowledge control, traceable reasoning, and protection against unsupported or overly general outputs.

To address these limitations, recent studies in defect detection and diagnosis have increasingly focused on linking large language models with structured knowledge representations. For example, Ma et al. proposed a knowledge graph-enhanced large language model pipeline for fault diagnostic reasoning and maintenance decision support, showing how structured domain knowledge can improve reasoning and support maintenance decisions in Industry 5.0 settings [15]. Zhou et al. (2024) developed CausalKGPT for root-cause analysis of quality problems in aerospace product manufacturing, leveraging a causal knowledge graph to enhance its reasoning capabilities [16]. Liu et al. (2024) presented a collaborative knowledge graph and large language model framework for fault diagnosis in aviation assembly, in which graph-structured knowledge is incorporated into the reasoning workflow through retrieval augmentation and model adaptation, without retraining the model [17]. These studies demonstrate that generative models based on and constrained by structured domain knowledge can significantly improve domain specificity and interpretability. However, comparable work remains limited in additive manufacturing defect analysis, especially in LPBF, where real-time decision support requires hierarchical defect organization, interpretation of representative defect micrographs, and defect-specific mitigation guidance.

Existing LPBF-focused studies further illustrate this gap. LLM-based approaches like AdditiveLLM have demonstrated the ability to predict a limited number of defect regimes, such as keyhole defect, lack of fusion, and balling, using structured inputs and natural language prompts to achieve robust classification and process parameter guidance [18]. Retrieval-augmented generation systems such as AMGPT support contextual querying of the additive manufacturing literature [19], while prior ontology-based work has provided a formal representation of LPBF defect-cause relationships [1]. Vision-based approaches have also shown strong capability for detecting and classifying hidden internal defects in additively manufactured parts [10]. However, these approaches generally address isolated capabilities, such as prediction, retrieval, representation, or detection, without integrating structured knowledge, flexible reasoning, and explainable interaction into a unified decision-support framework. LPBF-focused LLMs are also limited in defect coverage and lack explicit representation of defect hierarchies, inter-defect relationships, literature-supported explanation, and interactive support for defect investigation. A brief analysis of recent studies on LPBF defect analysis approaches and methods, as well as advanced LLM-driven diagnostics in broader manufacturing sectors, summarized in Table 1, provides an overview of the scope for implementing LLM agents in additive manufacturing defect identification. Prior work reveals a persistent gap between structured knowledge of defects, interactive reasoning, and actionable decision support for LPBF processes.

To address this gap, this work develops a knowledge-driven decision-support prototype for LPBF defect analysis and mitigation guidance by integrating an ontology-guided hierarchical knowledge base with LLM-based reasoning and interactive query handling within a unified workflow. The proposed framework is built on a structured representation of LPBF defect knowledge derived from published taxonomies and is implemented to support interactive defect exploration. The resulting knowledge base contains 27 LPBF defect types organized into hierarchical categories, together with curated causal information and mitigation guidance drawn from the literature. Within the proposed workflow, the large language model serves as the reasoning and interaction layer, enabling defect exploration via natural language queries and multimodal inputs. At the same time, the knowledge base constrains the analysis through curated defect categories, causal mappings, and, where available, material-specific mitigation guidance. Rather than functioning as a black-box predictor, the proposed LPBF defect agent is intended as a human-in-the-loop decision-support prototype that supports interactive defect exploration, literature-supported explanation, and hypothesis-guided interpretation of representative defect micrographs.

**Table 1**: Comparison of system capabilities across recent decision support frameworks for defect identification and diagnosis

| Ref. | Primary contribution | System architecture and features | | | | | |
|---|---|---|---|---|---|---|---|
| | | Conversational user interface | Embedded ontology or knowledge graph | Dynamic literature retrieval | Image analysis | Causal explanation | Rule-based mitigation guidance |
| [2] | Machine learning-based process and parameter optimization | | | | | | ✓ |
| [3] | In-situ layer-wise defect monitoring using a convolutional neural network | | | | ✓ | | |
| [4] | Predictive modeling of parameter–defect relations via knowledge graphs | | ✓ | | | ✓ | |
| [1] | Standardized semantic representation of defect-cause relations using the web ontology language | | ✓ | | | ✓ | ✓ |
| [8] | Document-level question answering for defect analysis using an LLM and a vector database | ✓ | | ✓ | | | |
| [18] | Predicting potential defect regimes by using the process parameter defect dataset | ✓ | | | | ✓ | |
| [15] | Fault diagnostic reasoning and maintenance using KG-enhanced LLMs | ✓ | ✓ | | | ✓ | |
| [16] | Aerospace fault cause analysis using causal knowledge-enhanced LLMs | ✓ | ✓ | | | ✓ | |
| [17] | Aviation assembly fault diagnosis via Prefix-Tuned Collaborative KG-LLM | ✓ | ✓ | ✓ | | ✓ | |
| This paper | Integration of multimodal visual diagnostics with deterministic, ontology-bounded mitigation through retrieval augmented generation | ✓ | ✓ | ✓ | ✓ | ✓ | ✓ |

The proposed knowledge-driven LLM-based decision-support system performs four related functions: (1) interactive and image-guided exploration of LPBF defects for analysis and decision making, (2) identification of major defect causes and suggestion of mitigation strategies, (3) defect-specific reasoning supported by hierarchical knowledge representation and fuzzy natural language query handling, and (4) interpretation of representative defect micrographs within a unified defect analysis workflow. Together, these functions are intended to support more accessible and structured defect exploration for LPBF users who may not possess highly specialized domain expertise.

The primary contribution of this work is the integration of four capabilities into a single LPBF-specific prototype framework. First, the proposed multimodal large language model-based agent combines a structured hierarchical defect knowledge base with contextual information from the additive manufacturing literature and optional evidence retrieval from external sources. Second, the framework links defect knowledge to material and process-specific mitigation guidance through curated mappings derived from published studies. Third, it supports interactive defect investigation through fuzzy natural-language query handling, stepwise clarification of broad defect categories into specific defect types, and literature-supported explanations. Fourth, it incorporates a multimodal image-analysis component that enables hypothesis-guided interpretation of representative defect micrographs through a heuristic semantic alignment score intended to improve interpretability. In contrast to approaches that rely on large, dynamically retrieved knowledge graphs, the present framework uses an ontology-guided hierarchical knowledge base to constrain the reasoning workflow in a computationally lightweight multimodal setting. Within the current prototype scope, this design supports defect-specific and context-aware analysis while preserving transparency in how defect information and mitigation guidance are presented.

The rest of the article is organized as follows: Section 2 provides a detailed description of the methodological framework for the ontology-integrated LLM-based decision support system, including hierarchical representation of defect knowledge, LLM-based reasoning and interactive query handling, and multimodal analysis of defect micrographs. Section 3 presents preliminary performance evaluation results for the proposed framework within the scope of the current study. Finally, Section 4 concludes the paper by summarizing the contribution, clarifying current limitations, and outlining directions for future work.

## 2. Methodology

This section presents the methodological framework of the proposed knowledge-driven LLM-based decision support system to aid LPBF defect analysis and mitigation. The framework is designed to support structured and interpretable decision support by combining a hierarchical LPBF defect knowledge base, large language model-based reasoning and query handling, optional literature-supported contextual augmentation, and multimodal image analysis within a unified workflow. The proposed methodology uses curated defect knowledge to support defect exploration, causal explanation, and mitigation guidance, rather than relying solely on unconstrained generative reasoning.

As illustrated in Figure 1, the proposed methodology consists of four interrelated stages: (1) construction of an ontology-guided hierarchical LPBF defect knowledge base, (2) query interpretation and defect-specific reasoning through fuzzy natural language handling and hierarchical defect clarification, (3) multimodal defect analysis with optional external evidence and knowledge augmentation, and (4) interactive deployment of the integrated workflow for prototype-level decision support. The four stages are briefly described below:

1. **Knowledge base construction**. At the first stage, a structured LPBF defect knowledge base was constructed from published literature on LPBF defects and their causes. The resulting representation contains 27 defect types organized into four broad categories: global structural defects, local structural defects, surface defects, and material defects. For each defect, the knowledge base stores hierarchical category information, along with curated causal factors and mitigation guidance when supporting literature was available. This representation provides the structured backbone for subsequent defect exploration and reasoning.
2. **Query interpretation and reasoning.** In the second phase, a reasoning layer was developed to interpret natural language user queries and map them to the appropriate defect information stored in the structured knowledge base. A "Smart Search" algorithm was developed using direct keyword matching and diffilib-based fuzzy matching to resolve misspelled or incomplete inputs, such as mapping “porsity” to "porosity". Additionally, smart filtering logic was implemented to handle broad queries. If a user searches for a general category like 'Cracking', the system identifies the ambiguity, and it displays a detailed list of defect types to ensure the final parameter recommendations are relevant to the defect.

3. **Multimodal analysis and contextual augmentation**. In the third phase, two complementary modules were integrated to extend the diagnostic workflow beyond text-based querying. The first one is an external evidence retrieval module based on SerpApi that can access scholarly and web-based sources to provide supplementary contextual information. The second one is a multimodal image-analysis module based on a foundation model that supports hypothesis-guided interpretation of representative LPBF defect micrographs. In the proposed framework, external knowledge retrieval is treated as an auxiliary evidence layer for the curated internal knowledge base. Additionally, the image-analysis module is intended to support interpretable defect assessment through a heuristic semantic alignment score.

4. **Interactive deployment and decision support**. In the final phase, the integrated workflow was deployed as an interactive prototype in Google Colab, supporting both menu-based navigation and free-text inquiry. Through this user-friendly interface, users ranging from researchers to shop-floor workers can explore defect categories, retrieve defect-specific causal and mitigation information, examine representative images and literature-supported context, and perform image-guided defect assessment within a single decision-support workflow. The present implementation should therefore be interpreted as a prototype framework for structured LPBF defect exploration, which can be extended to a validated industrial decision-support system.

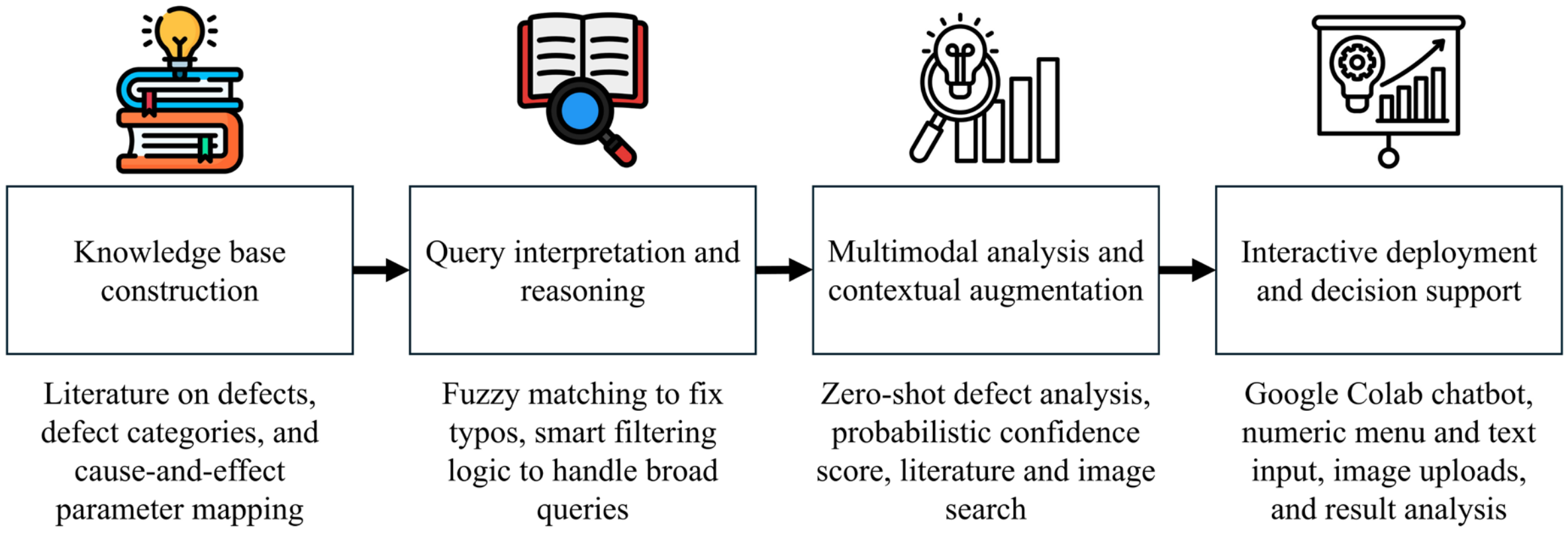

**Figure 1:** Methodological framework of the proposed knowledge-driven LLM-based decision-support system

These four steps collectively form an integrated reasoning pipeline in which structured defect knowledge serves as the primary basis for interpretation, while LLM-based reasoning, optional evidence retrieval from scholarly articles and the web, and multimodal image analysis extend the workflow’s accessibility and explanatory capability. The remainder of this section describes the system architecture in Section 2.1, the design and construction of the LPBF defect ontology and knowledge base in Sections 2.2 and 2.3, the natural-language reasoning and interaction mechanisms in Sections 2.4 and 2.5, the interpretable image-analysis components in Section 2.6, and the integrated workflow in Section 2.7.

### 2.1 System overview

The proposed LBBF defect agent is an interactive decision-support system implemented in Python to support LPBF defect identification, classification, and mitigation. The framework integrates two complementary intelligence components. The first is a knowledge-driven layer that leverages a hierarchical defect knowledge base, rule-based knowledge traversal, fuzzy query processing, and curated literature-derived mappings to support defect exploration and provide mitigation guidance. The second is a multimodal reasoning layer that supports image-based interpretation of representative defect micrographs using a foundation model within a hypothesis-guided workflow.

Within this integrated framework, the knowledge-driven layer serves as the primary source of structured defect information. It provides defect categorization, causal information, and mitigation guidance, all curated from the literature. The multimodal layer complements this functionality by supporting image-guided defect assessment when a user wishes to evaluate a representative defect micrograph. In addition, the framework can access scholarly and web-based sources through a retrieval module to provide supplementary contextual information and recent literature. It leverages the SerpApi–Google Search and Google Scholar interfaces, ensuring that recent studies are incorporated into the results [20].

To provide diagnostic accountability, the design imposes strict source verification on all outputs via a conditional routing protocol. The LLM is restricted to visual anomaly classification. On the other hand, internal fuzzy search algorithms map these visual classifications to formalized system nodes. For the next mitigation technique, the source is strictly defined by the material profile. For formally defined materials (e.g., IN625), the mitigation parameters are derived only from the deterministic Curated Knowledge Base (JSON application ontology). However, for unmapped or novel alloys, the system uses a Web Retrieval pipeline that summarizes generalized heuristics from the specific academic literature. All outputs are properly identified with their source origin (Ontology vs External Retrieval) for complete transparency to the end-user.

While the dynamic literature retrieval ensures the system has access to the latest academic insights, dynamically retrieved sources may contain conflicting or out-of-context process recommendations. Hence, external retrieval is treated as an auxiliary evidence layer rather than as a replacement for the curated internal knowledge base. To preserve consistency, the framework implements an ontology-first conflict-resolution strategy: when externally retrieved information is inconsistent with the encoded internal knowledge base, the curated defect definitions and material-specific mitigation mappings stored in the internal representation take precedence. This design allows external evidence to enrich explanation and context without overriding the structured defect knowledge embedded in the current prototype agent.

The prototype was deployed in Google Colab as an interactive command-line chatbot that supports both menu-based navigation and free-text user input. Through this interface, users can browse the defect hierarchy, retrieve defect-specific explanations, examine literature-supported context, and upload representative images for hypothesis-guided assessment. Figure 2 presents the initial interface and system options of the LPBF defect agent. Although the primary objective of the developed agent is to help shop-floor workers, researchers, and quality specialists with defect analysis and parameter suggestions, the agent also includes an automated reporting module for recent output to improve visualization.

For consistency across evaluations, the system's inference engine was mathematically constrained to ensure complete reproducibility and eliminate the stochastic variability inherent to large language models. Specifically, the API generation configuration for the core model was strictly locked to a temperature of 0.0, a top-p of 0.1, and a top-k of 1. These hyperparameter constraints turn off probabilistic sampling. Thus, it forces the model to select only the absolute highest-probability tokens. Therefore, the agent behaves more like a highly deterministic classification engine than a creative text generator. This restriction guarantees that identical microscopic image inputs yield identical semantic alignment percentages and mitigation strategies across independent research trials.

```
LPBF Defect Agent is (Smart NLP Search & Image Analysis) Ready!

📘 Available Options
─────────────────────────────
[1] → Show main defect types
[2] → List categories
[3] → Classify a defect (Supports fuzzy search)
[4] → Explore a defect (Numeric Menu)
[5] → Export Output (HTML & PNG)
[6] → 🔍 Analyze User Image (AI Vision)
─────────────────────────────
[0] → Back to Home
─────────────────────────────
```

**Figure 2:** Initial interface and system options of the LPBF defect agent

### 2.2 Ontology-guided knowledge base design

In this work, the term ontology refers to a structured, application-oriented representation of LPBF defect concepts and their relationships. More broadly, an ontology provides a formal semantic representation for organizing domain concepts and their interconnections in a structured, machine-interpretable form [21]. In additive manufacturing, such representations are useful for systematically organizing knowledge of defects, causal mechanisms, and interdependencies, where defect types and subtypes are defined as classes and linked by relationships that represent causal links and dependencies [22]. A structured representation is particularly important in LPBF because prior defect studies do not always maintain terminological consistency or hierarchical uniformity. Several earlier works have identified a wide range of defect modes, including porosity, cracking, surface roughness, and distortion, and have emphasized their relation to process parameters and part quality. At the same time, some studies have labeled mechanical outcomes, such as low strength or degraded mechanical properties, as defects, even though these outcomes are more appropriately interpreted as consequences of underlying physical defect mechanisms rather than as independent defect classes [23,24]. In other cases, broad defect categories and specific defect instances are presented at the same conceptual level, which makes cross-study comparison difficult. Terminological variation further contributes to ambiguity, as similar defect phenomena may be labeled as incomplete fusion holes [25], lack-of-fusion layer defect [23], and keyhole collapse [26]. These inconsistencies underscore the need for a more explicit, hierarchically organized representation of LPBF defect knowledge.

This need for a systematic framework led to the adoption of the LPBF defect taxonomy proposed by Hasan et al. (2023), which categorizes LPBF defects by formation mechanism and physical scale into four levels: global structural, local structural, surface, and material defects [1]. This classification provides clear boundaries between hierarchical layers and offers one of the most systematic conceptual frameworks available for LPBF defect modeling. Accordingly, this taxonomy was adopted as the foundation of the knowledge base developed in this work. To support mitigation guidance within the knowledge base, process parameter recommendations for Inconel 625 reported by Paraschiv et al. (2022) were also incorporated, where relevant supporting evidence was available [27].

The choice of an application-oriented hierarchical representation is also consistent with broader discussions of the practical use of ontology in manufacturing. Formal semantic interoperability in manufacturing is often associated with logic-based ontologies such as OWL and RDF, as well as query mechanisms such as SPARQL. Although such approaches offer significant interoperability benefits, they may be more rigid than necessary for an interactive prototype focused on defect exploration and decision support. Ameri et al. (2022) demonstrated that manufacturing knowledge integration may benefit from a composable stack of ontologies with varying levels of formality and coverage, rather than a single, monolithic standard [28]. They discussed the Industrial Ontology Foundry (IOF) model, which is a hub-and-spoke type. The core consists of basic reference ontologies, and the spokes are very specific "application ontologies" that meet specific software and operational needs. Viewed in that context, the present framework is best interpreted as a domain-specific, application-oriented knowledge structure designed to support LPBF defect reasoning within the scope of the current prototype.

Within this framework, the resulting representation consists of four top-level defect families as illustrated in Figure 3: (1) global structural defects, which are large-scale discontinuities that compromise structural integrity; (2) local structural defects, which include localized interruptions or imperfections in material continuity [29]; (3) surface defects, which manifest as imperfections on the final part surface or the active scan surface [30]; and (4) material defects, which arise from thermal stresses and affect microstructural integrity and mechanical performance [30]. The LPBF defect knowledge base was implemented as a hierarchical data structure using nested Python dictionaries. Each key represents a higher-level defect family, and each value contains either another dictionary or a list of specific defect instances. This hierarchical representation enables flexible, transparent modeling of multilevel defect relationships while remaining computationally lightweight and easily extensible.

The current iteration of the knowledge base includes 27 critical LPBF defects. This set should be interpreted as a representative prototype rather than as an exhaustive catalog of all possible LPBF defect manifestations. Even so, the representation is intended to be extendable. A primary advantage of the deterministic JSON architecture used in this work is that it inherently supports continuous validation and updating. As new metallurgical studies emerge or additional material-specific information becomes available, new defect entries, causal links, and parameter bounds can be incorporated into the hierarchical structure without requiring substantial modification of the overall workflow. In this way, the knowledge base can be updated by domain experts while preserving the core structure of the present prototype.

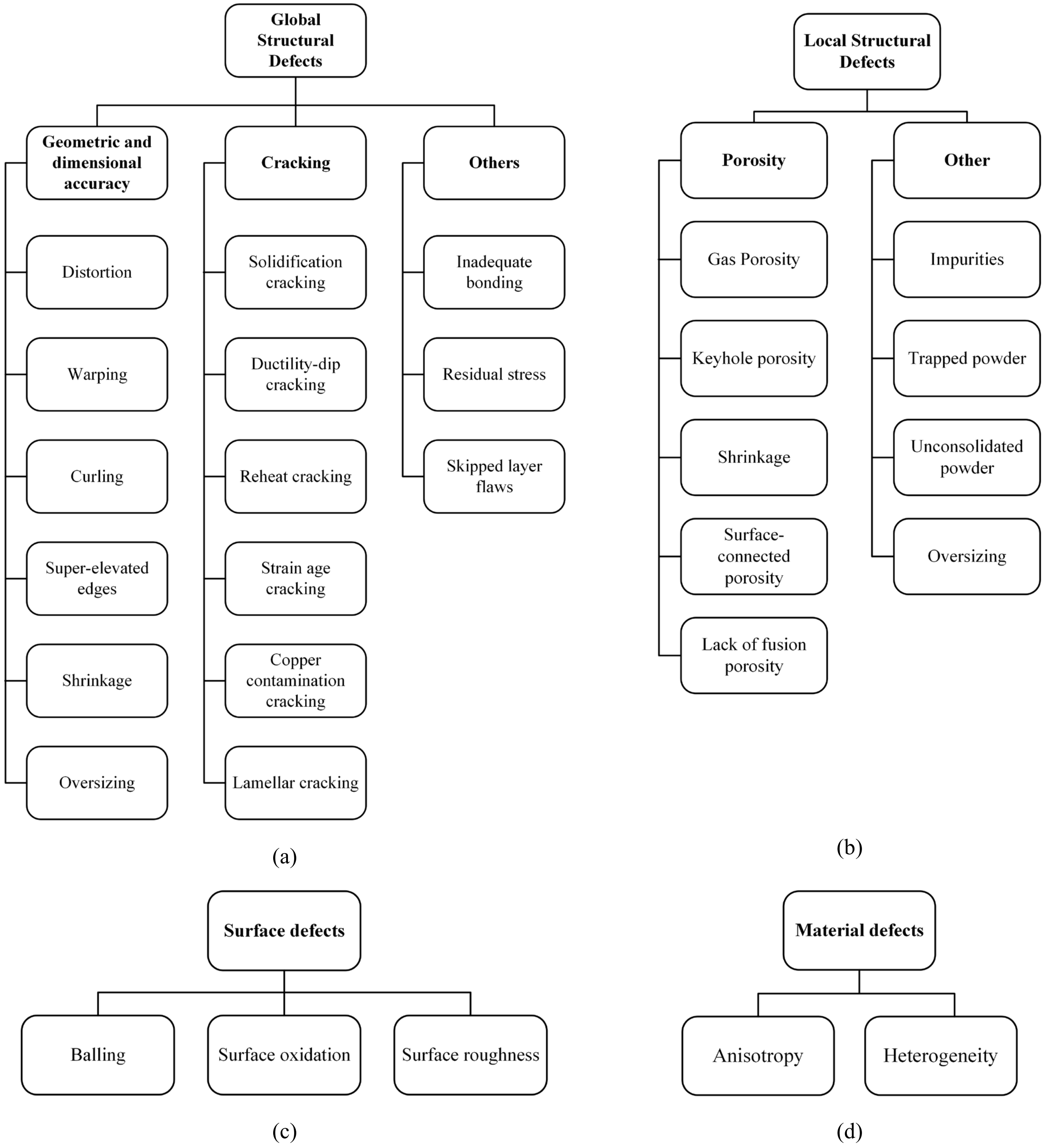

**Figure 3:** Categorization of high-level LPBF defects (a) global structural defects; (b) local structural defects; (c) surface defects; and (d) material defects

Because LPBF defects are organized across multiple levels of abstraction, explicit encoding of hierarchy is important for both reasoning and representation. Broad categories such as global and local structural defects encompass more specific subcategories, such as cracking or porosity, which in turn include distinct defect types. Representing these levels explicitly allows the agent to support both high-level defect exploration and more fine-grained defect interpretation during analysis. To operationalize this hierarchical structure, a recursive traversal algorithm was implemented to navigate the knowledge base from general categories to specific defect instances. The *Traverse_Defect_Categories* procedure serves as the entry point, iterating through the primary knowledge base keys. For each category, it calls a recursive function, *Traverse_Node*, which evaluates the structural depth of the current element and determines whether further traversal or defect-level reasoning is required. The pseudocode presented in Algorithm 1 outlines this traversal logic and illustrates how hierarchical ontology navigation is integrated into the agent's reasoning workflow.

Algorithm 1: Recursive traversal of the hierarchical defect ontology

| | |
|---|---|
| **Input** | *Defect_categories*: hierarchical dictionary of LPBF defects |
| **Output** | Printed traversal of all categories, subcategories, and defects |

```
1   Procedure Traverse_Defect_Categories (defect_categories)
2       for each key Category in defect_categories do
3           print (Category)
4           call Traverse_Defect_Node (defect_categories [Category], indent=1)
5       end for
6   end Procedure
7   Procedure Traverse_Defect_Node (node, indent)
8       if node is a List:
9           for each defect in the node do
10              print indentation (indent), defect
11          end for
12      else if node is a Dictionary:
13          for each key Subcategory in node do
14              print indentation (indent), Subcategory
15              call Traverse_Defect_Node (node [Subcategory], indent+1)
16          end for
17      end if
18  end Procedure
```

In addition to traversal, context-aware defect interpretation requires the system to recover the full hierarchical path associated with a queried defect. To support this functionality, the framework employs a recursive path-finding algorithm outlined in Algorithm 2 that accumulates parent category keys as it traverses the hierarchy. As a result, the system reconstructs the complete category path associated with a defect (e.g., local structural defects→ porosity), thereby preserving the semantic context required for accurate defect classification and exploration. The traversal follows a depth-first search strategy and continues until the target defect is located or all branches are exhausted. When the algorithm encounters a leaf node represented as a list, it checks whether the target defect is present and returns the accumulated path if a match is found. If the current node is a dictionary representing a defect category, the algorithm recursively explores each subnode while updating the traversal path. The default return of NULL indicates that the target defect does not exist in the current branch.

### 2.3 Knowledge base construction

A structured knowledge base was created to support defect-specific reasoning, causal explanation, and mitigation-oriented guidance within the proposed LPBF defect agent. This knowledge base was implemented using a hierarchical JSON representation to store defect entries, causal factors, and mitigation information in a transparent, computationally lightweight format. Within the present prototype, the knowledge base serves as the primary internal source of structured defect knowledge. It supports process-level reasoning by linking defects to documented causes and literature-derived mitigation guidance.

Each defect entry in the knowledge base is associated with three main types of information: its hierarchical position within the LPBF defect structure, its documented causal factors, and, where available, the corresponding mitigation guidance reported in the literature. For example, gas porosity is associated with causes such as entrapped gas and insufficient melting, together with parameter-level mitigation recommendations. By explicitly organizing these relations, the knowledge base enables the agent to retrieve defect-specific explanations and mitigation guidance in a structured, traceable way.

Algorithm 2: Recursive path-finding in the hierarchical LPBF defect knowledge base

| | |
|---|---|
| **Input** | Target Defect (T), Current Node (N), Current Path (P) |
| **Output** | Full Path List or NULL |

```
1   Procedure FIND_PATH (T, N, P)
2       if N is a List (Leaf Node) then
3           if T exists in N then
4               return P
5           end if
6       else if N is a Dictionary (Category) then
7           for each Key, SubNode in N do
8               New_Path←P+[key]
9               Result← FIND__PATH (T, SubNode, New_Path)
10              if Result ≠ NULL then
11                  return Result
12              end if
13          end for
14      end if
15  end Procedure
16  return NULL // Reached only if the target defect is not found in this branch
```

The knowledge base also supports limited causal reasoning by encoding sequential relationships among process conditions, intermediate physical mechanisms, and resulting defects. For example, an energy imbalance may contribute to melt-pool instability, which in turn can lead to balling and associated surface roughness. In the current prototype, such reasoning is not generated solely through unconstrained inference; instead, it is guided by curated relational mappings stored in the knowledge base. This design allows the large language model to generate natural language explanations that remain tied to structured defect knowledge rather than relying solely on open-ended text generation.

As illustrated in Figure 4, the knowledge base organizes entities such as process parameters, intermediate mechanisms, and defects through explicitly defined hierarchical and causal relationships. The knowledge base, therefore, functions as a constrained, domain-specific knowledge structure for LPBF defect analysis, supporting defect exploration, explanation, and mitigation-oriented reasoning within the scope of the current agent prototype. Where the literature provided sufficiently specific parameter guidance, those mappings were incorporated into the knowledge base for defect-specific mitigation recommendations. In the present implementation, such guidance is most explicitly curated for Inconel 625. Accordingly, the knowledge base should be interpreted as an application-oriented representation designed for prototype-level LPBF reasoning, and the associated mitigation recommendations should be understood as context-bound guidance derived from the literature rather than as a standalone solution applicable to all materials, machines, and LPBF process conditions.

### 2.4 Query interpretation and reasoning

The proposed LPBF defect agent supports two complementary modes of interaction: menu-based navigation and free-text natural-language querying. To interpret free-text input, the system implements a Smart Search mechanism that combines direct keyword matching with fuzzy string comparison. This design allows the agent to respond to both well-formed technical queries and noisier user input containing typographical errors, incomplete terminology, or broad defect categories.

The first stage of query interpretation uses substring matching to identify explicit defect terms embedded in longer user inputs. For example, a query such as "What is porosity?" is mapped directly to the corresponding defect category. If no exact match is found, the system uses fuzzy string matching via the difflib library as a fallback mechanism. Specifically, the

*get_close_matches()* function compares the user input against a flattened representation of the defect hierarchy [31]. This approach enables robust handling of typographical errors, such as mapping "porsity" to "porosity".

A further clarification step is applied when the interpreted query corresponds to a broad defect category rather than a specific defect instance. In such cases, the system prompts the user to refine the query by selecting from more specific child defects. For example, a query about porosity may be refined into gas porosity, keyhole porosity, lack of fusion porosity, or surface-connected porosity. This stepwise clarification process helps ensure that subsequent reasoning and mitigation guidance remain defect-specific and contextually appropriate.

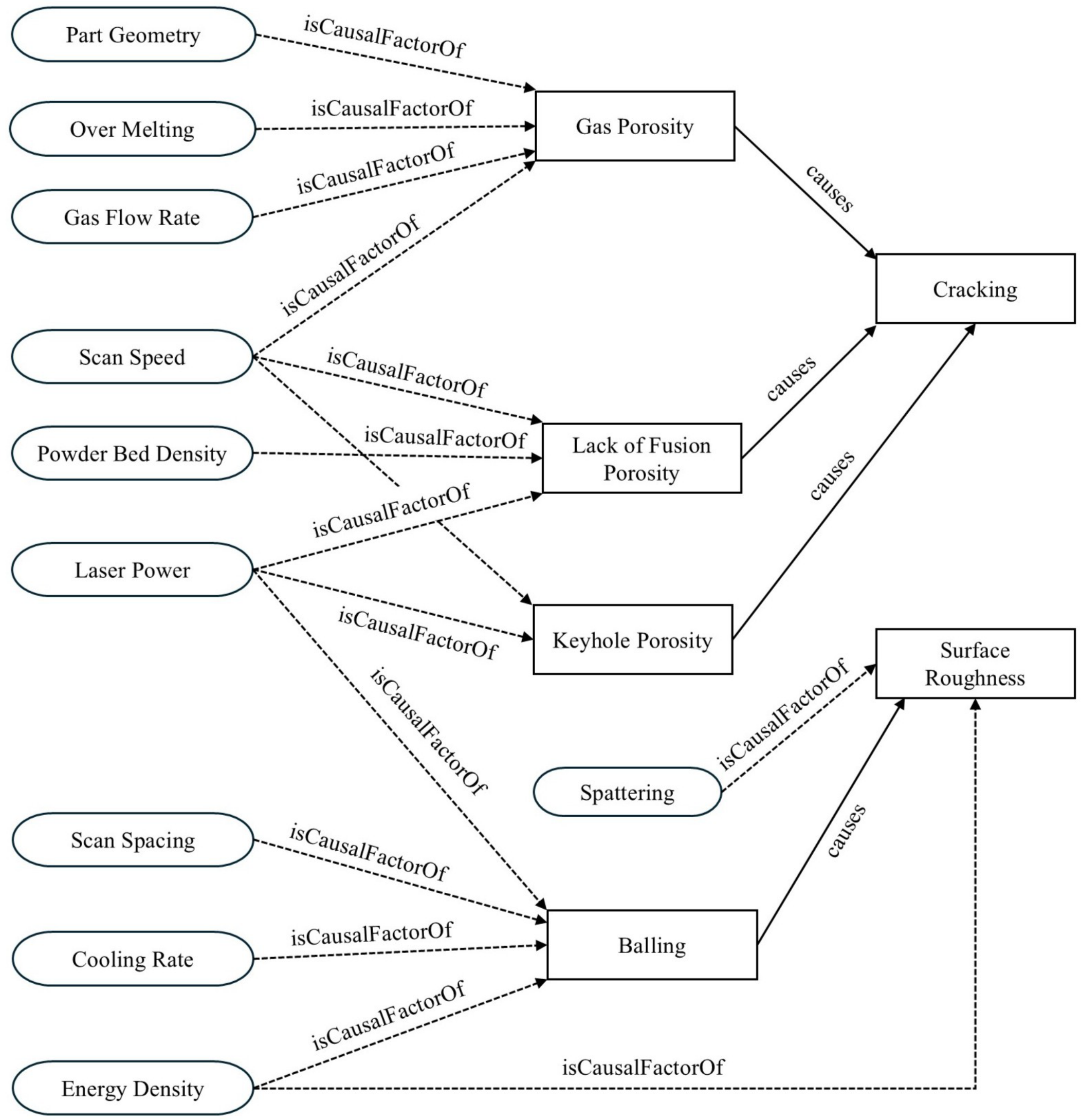

**Figure 4:** Graphical representation of the deterministic causal rules and relationships embedded in the agent's JSON knowledge base

### 2.5 Interactive deployment and decision support

Users can interact with the LPBF defect agent through the command-line interface illustrated in Figure 2. Upon selecting the option to explore a defect, the agent presents a structured list of available defect categories and types, and prompts the user to select the specific defect of interest. Each user selection triggers a three-stage interaction workflow: (1) ontology lookup, which retrieves the hierarchical category path associated with the selected defect; (2) knowledge-based querying, which displays causal explanations and process-parameter mitigation guidance from the knowledge base; and (3) optional multimodal augmentation, which integrates defect-related images and literature to support a thorough understanding. This interactive workflow enables users to move from broad defect exploration to more detailed and context-aware reasoning within a single interface. Rather than serving as a fully automated diagnostic system, the current workflow provides access to curated defect knowledge and supports systematic defect exploration and explanation through guided interaction and contextual augmentation.

```
Select a defect:
[1] Anisotropy
[2] Balling
[3] Copper contamination cracking
[4] Curling
[5] Distortion
[6] Ductility-dip cracking
[7] Formation of super-elevated edges
[8] Gas porosity
[9] Heterogeneity
[10] Impurities/Inclusions
[11] Inadequate bonding between layers
[12] Keyhole porosity
[13] Lack of fusion porosity
[14] Lamellar cracking/Delamination
[15] Oversizing
[16] Reheat and post weld heat treatment cracking
[17] Residual stress
[18] Shrinkage
[19] Skipped layer and stop/start flaws
[20] Solidification cracking
[21] Strain age cracking
[22] Surface oxidation
[23] Surface roughness
[24] Surface-connected porosity
[25] Trapped powder
[26] Unconsolidated powder
[27] Warping
```

**Figure 5:** Defect exploration interface of the proposed LLM agent

## 2.6 Semantic alignment evaluation for image analysis

The proposed LPBF defect agent includes a multimodal image-analysis module to support hypothesis-guided interpretation of representative defect micrographs. In the current implementation, visual feature extraction and multimodal reasoning are performed using the Gemini 2.5 Pro foundation model. The image-analysis workflow is guided by defect-specific descriptors and the structured LPBF knowledge base developed in this study, without any model fine-tuning on LPBF metallurgical data.

Unlike a conventional discriminative classifier, the image-analysis module is not designed to output a calibrated class probability. Rather, it produces a semantic alignment score reflecting the degree of correspondence between observed image features and defect-specific descriptors provided in the prompting context. This score is heuristic and interpretive. It is intended to support transparent comparison among candidate defect hypotheses and should not be interpreted as statistical confidence or expected classification accuracy. Typical discriminative classifiers use a *softmax* layer to estimate class probabilities [32]. By contrast, the present approach emphasizes interpretability by assessing whether observed visual characteristics are consistent with the expected morphology of a candidate defect. The image-analysis workflow proceeds through three stages, described below.

### *2.6.1 Visual feature encoding*

First, the input image is processed by the multimodal model and embedded in a high-dimensional latent vector that captures defect-related attributes such as void sphericity, edge sharpness, and the spatial distribution of pores. This latent encoding serves as the visual basis for the subsequent evaluation of defect-specific hypotheses. The present workflow relies on in-context guidance from structured defect descriptors rather than LPBF-specific fine-tuning of the underlying multimodal model.

#### *2.6.2 Hypothesis-guided prompting*

To guide defect evaluation, the agent incorporates hypothesis-conditioned prompting, in which the system explicitly queries whether the image corresponds to a suspected defect (e.g., "Are you suspecting 'X' defect?"). This conditioning mechanism directs the model's internal attention to the visual features most relevant to the underlying defect-formation mechanism, enabling targeted evaluation rather than automatic label assignment.

#### *2.6.3 Semantic alignment scoring*

In the third stage, the model’s response is interpreted as a semantic alignment score reflecting how closely the observed visual features match the defect descriptors. When the image exhibits a feature that strongly aligns with a candidate defect description, the corresponding semantic alignment score is higher. Conversely, when the observed morphology is inconsistent with the supplied defect characteristics, the score is lower. For example, strong alignment with keyhole porosity may be associated with deep, elongated, V-shaped cavities. In contrast, irregular voids with sharp boundaries and signs of incomplete melting may align more strongly with a lack of fusion. This interpretive workflow is intended to mirror the kind of feature-based comparison used in expert visual assessment, while keeping the output tied to structured defect descriptors.

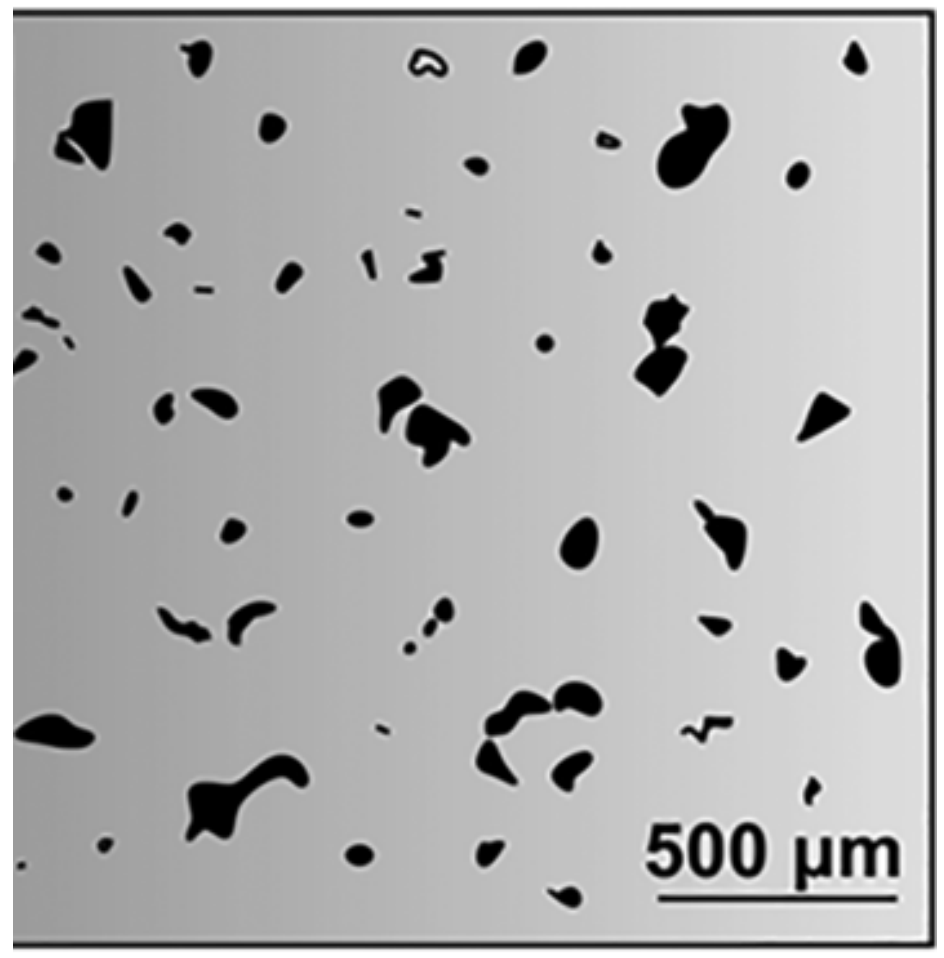

**Figure 6:** Representative microscopy image used to illustrate semantic alignment scoring

To illustrate this process, the microscopy image reported by Parsazadeh et al. [33], shown in Figure 6, is used as a representative example. The image exhibits an irregular, non-spherical void with sharp edges and unmelted powder particles. The multimodal model first forms a high-dimensional latent representation of the image that captures relevant morphological attributes, including irregular geometry, sharp void boundaries, internal powder presence, and the overall spatial architecture of the defect. When evaluated against the candidate defect hypotheses, the observed morphology aligns weakly with keyhole porosity but more strongly with lack of fusion. In this example, the agent assigns a semantic alignment score of 0.10 to keyhole porosity and 0.85 to lack of fusion, while also noting weak alignment with gas porosity (0.05) due to the presence of small spherical pores. This score pattern is qualitatively consistent with the observed morphology in the defect micrograph: the representative image does not exhibit the deep, vertically elongated cavity geometry typically associated with keyhole porosity. Instead, it shows irregular void morphology, sharp boundaries, and signs of unmelted powder, which are more consistent with lack of fusion. Table 2 summarizes the semantic alignment scores assigned to the candidate defect hypotheses and illustrates how the proposed workflow compares extracted visual evidence with defect-specific descriptors. For this representative image, the observed morphology aligns more closely with lack of fusion than with keyhole porosity, because the voids are irregular, have sharp boundaries, and are associated with unmelted powder particles.

**Table 2:** Semantic alignment scoring for LPBF defect assessment

| Feature dimension | Observed visual evidence | Lack of fusion descriptor [34] | Alignment with lack of fusion | Keyhole porosity descriptor [35] | Alignment with keyhole porosity |
|---|---|---|---|---|---|
| Morphology | Irregular, non-spherical void morphology | Irregular, sharp-edged voids | High | Smooth, rounded, or V-shaped cavities | Low |
| Edge profile | Sharp, jagged void boundaries | Sharp, angular boundaries | High | Smooth internal boundaries | Low |
| Interior content | Presence of unmelted powder particles | unmelted or partially melted powder | High | Gas or vapor-filled cavity with no entrapped powder | Low |
| Orientation relative to layers | Predominantly aligned with layer structure | Aligned with layers | High | More consistent with depth-oriented keyhole morphology | Low |
| Semantic confidence score | | | **0.85** | | **0.10** |

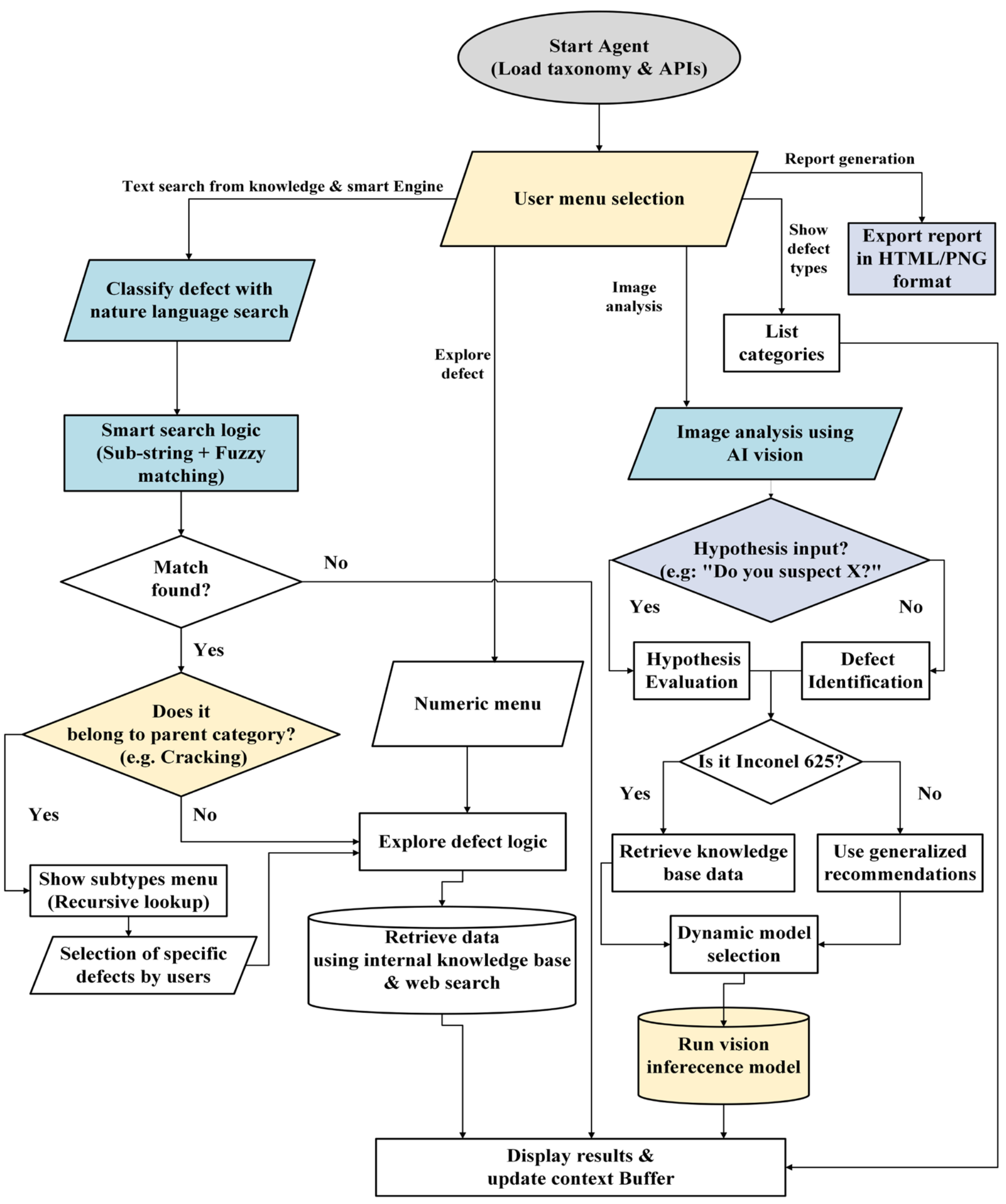

**Figure 7:** Operational flow diagram of the proposed LPBF defect agent

### 2.7 Integrated workflow of the LPBF defect agent

Figure 7 summarizes the operational logic of the proposed LPBF defect agent. Through the interface shown in Figure 2, the proposed agent integrates two complementary workflows within a unified framework: a text-based reasoning workflow for defect exploration and knowledge retrieval (options 1-4), and (2) an image-based assessment workflow for multimodal defect interpretation (option 6). At initialization, the system loads the hierarchical defect knowledge base together with the API configurations required to support interactive user input.

After initialization, the workflow branches based on the user's selected interface option. Options 1–4 invoke the text-based reasoning workflow, in which the Smart Search module processes natural-language input through substring matching and *difflib*-based fuzzy string matching. If the identified term corresponds to a broad defect category (such as cracking or porosity), the system applies recursive disambiguation and prompts the user to refine the selection to a more specific defect subtype. Once a specific defect is selected, the agent retrieves the corresponding causal information and mitigation guidance from the internal knowledge base. When requested, external scholarly retrieval through SerpApi and Google Scholar is used to provide supplementary contextual information and recent literature relevant to the queried defect. In the current prototype, this retrieval layer is used to augment explanation and context; it does not replace the internal knowledge base or override curated defect-specific mitigation mappings.

Option 6 activates the image-based assessment workflow. When the user uploads a representative defect micrograph, the workflow initiates a hypothesis-testing checkpoint and prompts the user to indicate whether a specific defect is suspected. This input governs the prompt construction strategy, determining whether the model performs general defect identification or targeted hypothesis evaluation. The inference layer uses a dynamic model selection mechanism to identify an appropriate multimodal foundation model variant, cycling between Flash and Pro versions to ensure reliability and robust visual interpretation across defect scenarios.

The output layer subsequently integrates results from both the text-based reasoning engine and the vision-based analysis module, presenting users with semantic alignment scores, supporting visual evidence, and process-level mitigation guidance. By integrating deterministic ontological reasoning with generative multimodal analysis, the workflow enables explainable, context-aware defect diagnosis and mitigation support, allowing users to seamlessly transition between knowledge exploration, image-based validation, and decision-support outputs.

To mitigate the inherent non-determinism and quality variation of dynamic web retrieval, the SerpApi module operates under a strict 'Ontology-First' hierarchy. The deterministic JSON knowledge schema serves as the primary ground truth for all IN625 defect mitigations. The dynamic search module is used exclusively for secondary context augmentation (e.g., retrieving recent literature to support the baseline recommendation). To ensure source quality, SerpApi queries are programmatically constrained to peer-reviewed databases. Furthermore, to address conflicting information, the LLM is instructed, through its system prompt, to discard any retrieved web parameters that violate the JSON schema's established boundaries. Finally, to support reproducibility despite the dynamic nature of web search, the agent maintains a reference audit trail, explicitly logging the titles and sources of any retrieved literature used in its final output.

At present, material-specific mitigation recommendations are limited to IN625, as the knowledge base contains curated recommendations for only that material. For other materials, the system relies on the LLM for broader reasoning support. Material selection is therefore included as an input step when the defect-analysis option is triggered. Additional material-specific information can be incorporated into the ontology knowledge base in future work.

## 3. Results and Discussion

This section presents experimental results and demonstrates the proposed ontology-integrated LLM agent for LPBF defect analysis, focusing on its ability to support structured defect exploration, user interaction, and explainable image-based reasoning. Unlike general-purpose LLMs, which operate without explicit process constraints, the proposed agent supports LPBF-specific reasoning through a structured internal knowledge base. This design constrains the model's response space to metallurgically meaningful interpretations and enables more structured defect exploration, explanation, and mitigation guidance within the scope of the current prototype.

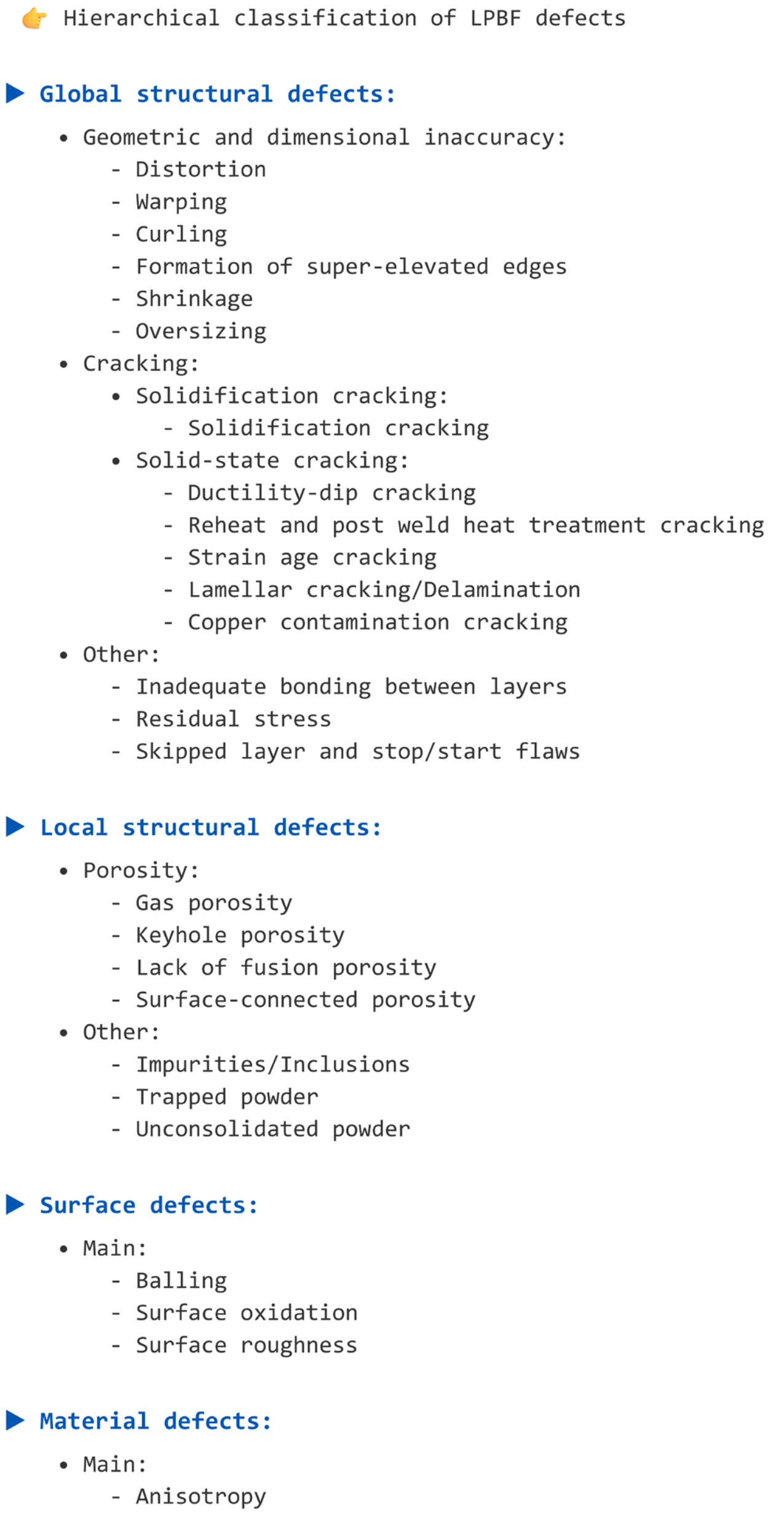

**Figure 8:** Hierarchical representation of LPBF defect knowledge used in the proposed agent

### 3.1 Structured representation of LPBF defect knowledge

As depicted in Figure 8, the proposed LLM agent enables structured representation and navigation of LPBF defect knowledge derived from a curated ontology. The agent presents defect information from a trusted internal knowledge base while retaining the ability to augment it with external scholarly sources upon request. This combination allows users to access both stable, validated internal defect knowledge and supplementary, literature-supported context, thereby supporting informed defect exploration without relying solely on unstructured web knowledge retrieval.

### 3.2 Interactive defect exploration

The agent can explore defects in two ways. It can accept both numeric input from the menu and manually entered text (e.g., "Explore gas porosity") from the user. An example interaction is presented in Figure 10. For numeric input, the system operates as a multi-stage defect exploration system that can explore defects by displaying knowledge base information, as well as images and summaries from scholarly articles with references. The step-by-step numeric defect exploration process is illustrated in Figure 10, where the user first selects "Explore a defect" from the menu. For example, if the user wants to explore the "Balling" defect, they can provide "2" as an input. The agent will then provide the relevant information from the knowledge base. Upon a user's request, the agent can perform live searches with SerpAPI to identify defects related to LPBF and summarize the most recent publications relevant to the defect. Additionally, the agent searches for images based on the user's input. This interaction demonstrates how the agent supports both guided menu-based navigation and free-text querying, allowing users to progress from broad defect exploration to more detailed, context-aware understanding while preserving control over the level of information requested.

### 3.3 Handling of fuzzy and ambiguous user queries

To improve usability for non-experts and for realistic use cases, the developed AI agent supports fuzzy matching of defect keywords. When a user inputs a misspelled or incomplete defect name, such as "porsity" instead of "porosity", the system automatically performs similarity matching using *difflib.get_close_matche*s and suggests the most likely intended defect (for example, "Did you mean 'Porosity'?"), while allowing the user to confirm or revise the match before proceeding. This approach enables the agent to handle noisy input and improves accessibility for users unfamiliar with exact technical terminology. Furthermore, common parent categories such as "porosity" or "cracking" are disambiguated by follow-up prompts that allow the user to select subtypes, such as "keyhole porosity", "gas porosity". An example interaction is shown in Figure 9. This recursive mechanism helps ensure that downstream reasoning and recommendations remain defect-specific and contextually appropriate.

```
--- Classification Chat Log ---
You: crackng
❓ Interpreted as 'Cracking'. Proceed? (yes/no)
You: yes
🔍 Multiple types of 'Cracking':
[1] Solidification cracking
[2] Ductility-dip cracking
[3] Reheat and post weld heat treatment cracking
[4] Strain age cracking
[5] Lamellar cracking/Delamination
[6] Copper contamination cracking
👉 Enter the number of your choice:
You: 1
```

**Figure 9:** Fuzzy-query handling for misspelled or ambiguous LPBF defect input

### 3.4 Dynamic retrieval of supplementary knowledge

A significant capability of the proposed LPBF defect agent is its dynamic knowledge retrieval mechanism. While the agent is built on a curated internal knowledge base of 27 LPBF defect types and associated process parameters, it provides access to supplementary information from external scholarly and web-based sources when requested. This retrieval-augmented approach allows the agent to provide recent literature context, representative images, and additional explanatory material related to the queried defect.

In the current prototype, the internal knowledge base remains the primary source of structured defect information, including the hierarchical defect organization and curated causal and mitigation mappings. External retrieval is used as an auxiliary layer to enrich explanation and context; it does not replace the internal knowledge base or override curated defect-specific mitigation mappings. This distinction is important because dynamically retrieved sources may vary in relevance, specificity, and consistency with the encoded knowledge base.

🔍 **Exploring: Balling**

Category: Surface defects → Main
Causes: High scan speed, Low melt pool temperature
Optimization Parameters:
- Laser Power: Increase to stabilize melt pool
- Scan Speed: Reduce at low power
- Layer Thickness: Use thinner layers (30–50 μm)
- Oxygen Level: Keep $O_2$ < 0.1 %

Note: Unstable melt pool wetting → balling.
--- Additional fetched resources ---

🖼️ **Image search for 'Balling':**

1. Balling effect caused by high laser power (adapted from [18, 56, 57]) | Download Scientific Diagram
   Source: **https://www.researchgate.net/figure/Balling-effect-caused-by-high-laser-power-adapted-from-18-56-57_fig3_342050556**

🌐 **Web search for 'Balling':**

1. A simple scaling model for balling defect formation during ...
   **https://www.sciencedirect.com/science/article/pii/S2214860423000441**
   We propose an outstandingly simple thermal scaling model for predicting the threshold from balling mode to conduction mode in laser powder bed fusion.
2. Balling Effect in LPBF: Causes, Impact, and How to Prevent It
   **https://insidemetaladditivemanufacturing.com/2025/02/28/deep-dive-understanding-and-preventing-the-balling-effect-in-laser-powder-bed-fusion-lpbf/**
   Balling occurs when the melt pool becomes unstable, causing the molten metal to form small droplets instead of a continuous track.

🎓 **Scholar search for 'Balling':**

1. A simple scaling model for balling defect formation during laser powder bed fusion
   … from balling mode to conduction mode in laser powder bed fusion. The resulting balling … number which combines the material properties, the powder size and the pre-heating of the …...
2. Analytical prediction of balling, lack-of-fusion and keyholing thresholds in powder bed fusion
   … This study proposes analytical methods to predict the defect generation in laser powder bed fusion additive manufacturing. The occurrence of lack-of-fusion, balling and keyholing …...
3. Numerical investigation of balling defects in laser-based powder bed fusion of metals with Inconel 718
   … We find a similar balling behavior yet different ball size at different laser power settings. We … to quantify balling processes and to improve the understanding of balling mechanisms. …...

**--- Consolidated Summary (AI Synthesized) ---**

Balling represents a critical defect mode in laser powder bed fusion (LPBF) additive manufacturing, exhibiting a distinct melt pool behavior from the desired conduction mode. Analytical methods are actively being developed to predict its occurrence, alongside other common defects such as lack-of-fusion and keyholing. The onset and characteristics of balling are influenced by a complex interplay of material properties, powder size, and pre-heating conditions, frequently consolidated within a specific dimensionless number. While studies indicate similar balling behavior under varying conditions, the resultant ball size is notably dependent on laser power settings. Current research endeavors are focused on quantifying balling processes and improving the fundamental understanding of its underlying mechanisms.

❓ *Would you like to see causal relationships expanded for 'Balling'? (yes/no): yes*

Factors leading to Balling:
- Energy density → Balling
- Laser power → Balling
- Scan spacing → Balling
- Cooling rate → Balling

Balling can lead to:
- Balling → Surface roughness

**Figure 10** Step-by-step menu-based workflow for interactive defect exploration

### 3.5 Demonstration of causal reasoning capability

In scenarios where causal relationships are explicitly encoded in the knowledge base, the proposed agent can retrieve and explain the corresponding cause-and-effect pathways, as illustrated in Figure 11. This capability extends the agent beyond descriptive defect identification by connecting selected defects to documented process conditions and intermediate mechanisms. In the current prototype, causal reasoning is limited to well-established defect relationships reported in the literature. Future work will focus on extending this capability to broader, indirect, and multi-defect causal interactions.

```
❓ Would you like to see causal relationships expanded for 'Balling'? (yes/no): yes
Factors leading to Balling:
 • Energy density → Balling
 • Laser power → Balling
 • Scan spacing → Balling
 • Cooling rate → Balling
Balling can lead to:
 • Balling → Surface roughness
```

**Figure 11:** Demonstration of encoded causal reasoning for LPBF defect analysis

### 3.6 Example output of AI-assisted defect assessment

Figure 12 illustrates an example of AI-assisted defect-assessment output generated by the proposed agent for a representative LPBF defect micrograph. The agent identifies candidate defect mechanisms, reports semantic alignment scores, highlights relevant visual features, and suggests mitigation-oriented guidance when corresponding information is available in the internal knowledge base or through literature-supported context. For the representative example shown in Figure 12, the vertically elongated cavity penetrating from the surface (indicated in green) is associated with keyhole porosity, whereas the smaller, scattered voids (indicated in red) are associated with irregular gas porosity.

The associated semantic alignment scores indicate the degree of correspondence between the observed image morphology and the defect-specific descriptors, rather than calibrated classification probabilities. For this sample defect micrograph, the agent assigns a semantic alignment score of 0.90 to keyhole porosity and 0.70 to irregular gas porosity. These values reflect the relative correspondence between the visual evidence, including cavity geometry and pore distribution, and the morphological descriptions associated with the candidate defect types. Unlike conventional convolutional neural network classifiers, which typically map image features directly to discrete class labels or probability-like outputs, the proposed image-assessment workflow uses descriptor-based semantic alignment. This approach supports interpretability by linking candidate defect outputs to observable features in the defect micrograph and structured defect knowledge.

The example also illustrates how the agent connects image-assisted defect assessment with guidance aimed at defect mitigation. The current defect agent prototype focuses primarily on IN625 because the material-specific mitigation mappings encoded in the knowledge base were curated for this alloy. However, the framework is not inherently limited to IN625; it can be extended to other LPBF material systems by incorporating material-specific defect mechanisms, process windows, and mitigation guidance into the knowledge base. For materials not yet encoded in this way, the agent relies more heavily on generalized large-language model reasoning and literature-supported context; therefore, such outputs should be interpreted cautiously, as equivalent material-specific mappings have not yet been curated. This example demonstrates how the prototype integrates visual interpretation, structured defect knowledge, and mitigation-oriented reasoning within a single LPBF defect-assessment workflow.

## 4. Empirical Evaluation and Comparative Analysis

This section evaluates the proposed LPBF defect agent through three complementary analyses. First, a qualitative comparison is conducted to examine how the proposed agent differs from general-purpose vision-language models in interpreting the same representative LPBF defect micrograph. Second, an ablation study is used to assess the contribution of the main architectural components, including the base language model, dynamic retrieval, and ontology-guided knowledge base. Third, inter-rater reliability analysis is used to examine agreement between model outputs and expert reference labels. Together, these analyses evaluate the prototype's diagnostic behavior, component contributions, and agreement with expert interpretation within the scope of the available literature-derived evaluation dataset.

🔍 **AI Analysis for: defect_3.png**

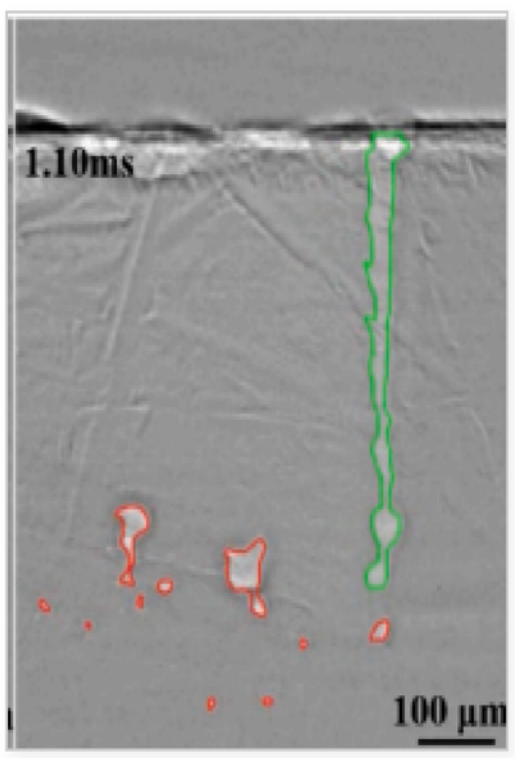

**--- Defect Analysis ---**

```
1.  **Keyhole Porosity**: 90% Probability
    *   **Visual Evidence**: A prominent, deep, narrow, and elongated void (highlighted in green) that
originates directly from the top surface and extends significantly into the material. The internal shape
is somewhat irregular, suggesting an unstable formation process rather than a perfectly stable void.
    *   **Reasoning**: This morphology is highly characteristic of keyhole porosity. It forms when the
laser power is too high, creating an excessively deep vapor cavity (keyhole) in the melt pool. If this
keyhole becomes unstable or collapses before the molten metal can fill the void, gas is trapped, forming
such a defect.
2.  **Gas Porosity (Irregular)**: 70% Probability
    *   **Visual Evidence**: Several smaller, irregularly shaped voids (highlighted in red) are scattered
within the material, beneath the surface. They are not perfectly spherical but also not angular,
suggesting trapped gas rather than lack of fusion between layers.
    *   **Reasoning**: While classic gas porosity is typically spherical, these irregular shapes can
arise from gas entrapment under specific melt pool dynamics, especially if the melt pool is unstable due
to factors that also cause keyhole porosity. The gas source could be dissolved gases in the powder,
shielding gas, or partial keyhole collapse. Their size and relative isolation are consistent with trapped
gas pockets.
```

**--- Correction Strategy ---**

```
**Recommendation**: The primary defect identified is keyhole porosity, which indicates an overly
aggressive or unstable melt pool. The smaller irregular pores could also stem from similar root causes
related to melt pool dynamics. To mitigate these issues, the following parameter adjustments are
recommended:
1.  **Reduce Laser Power**: Lowering the laser power will decrease the energy density, preventing the
formation of an excessively deep and unstable keyhole.
2.  **Increase Scan Speed**: A higher scan speed reduces the interaction time of the laser with the
material at any given point, effectively reducing the energy input per unit area and thereby suppressing
keyhole formation.
3.  **Optimize Focus Offset**: Adjusting the laser focus offset to a slightly defocused state can broaden
the melt pool and reduce the peak power density, making keyhole formation less likely while improving
melt pool stability.
4.  **Improve Gas Flow/Purity**: Ensure the inert shielding gas flow is optimized and the gas purity is
high to minimize the introduction of exogenous gases that could contribute to porosity.
```

**Figure 12:** Example defect-assessment output showing candidate defect mechanisms, visual annotations, semantic alignment scores, and mitigation-oriented guidance

### 4.1 Qualitative comparison with vision language models

To examine how structured LPBF defect knowledge influences model output, a qualitative comparison was conducted between the proposed LPBF defect agent and two general-purpose vision language models using the same representative LPBF defect micrograph and diagnostic prompt. Note that the purpose of this comparison was not to establish a statistically comprehensive benchmark, but to examine differences in visual interpretation, defect assessment, and mitigation guidance under a common input condition. The diagnostic prompt was formulated as follows: *"Analyze this LPBF micrograph showing a specific defect. Identify the defect type, root cause, and recommend mitigating parameter adjustments for IN625"*. The representative micrograph used for this comparison was adapted from [36] and shown in Figure 13. The comparative outputs from the vision language models and the proposed defect agent are summarized in Table 3.

**Table 3:** Qualitative comparison of the proposed LPBF defect agent with general-purpose vision-language models

| AI system | Visual interpretation | Defect assessment | Mitigation guidance | Interpretation of output |
|---|---|---|---|---|
| ChatGPT-5.3 | From the LPBF micrograph, the defect morphology is characterized by irregular, jagged, non-spherical dark voids distributed along what appear to be melt track boundaries rather than perfectly round pores | Lack of fusion porosity | 1. Increase energy input (laser power, scan speed)<br>2. Improve track overlap<br>3. Reduce layer thickness. Thinner layers promote full remelting and bonding. | Identified the primary defect consistently with the reference interpretation and provided general mitigation guidance |
| Gemini 3.1 Pro | Looking at this micrograph, the defect shown is classic Lack of Fusion (LoF) porosity | Lack of fusion porosity | 1. Decrease scan speed<br>2. Increase laser power<br>3. Reduce hatch spacing | Identified the primary defect consistently with the reference interpretation and provided general mitigation guidance |
| Proposed LPBF defect agent | The image shows numerous dark, irregularly shaped, and often elongated voids distributed throughout the material. Many of these pores are relatively large and non-spherical, resembling unfused regions between melt pools or layers. | Lack of fusion porosity; semantic alignment score = 0.90 | Recommendation: To mitigate the lack of fusion porosity in IN625, the following deterministic parameter changes are recommended based on the ontology:<br>Laser Power: Increase laser power, specifically aiming for > 150–200 W, and ideally within the ~250–300 W range for IN625, to ensure full melting and better melt pool stability.<br>Scan Speed: Decrease scan speed, particularly if it is currently above 900 mm/s, to provide more interaction time for the laser and enhance melting and fusion.<br>Volumetric Energy Density (VED): Ensure the VED is maintained within the range of 65–90 J/mm³ by adjusting laser power, scan speed, layer thickness, and hatch spacing. This range is critical for achieving high density (≥ 99.5%) in IN625. | Identified the primary defect consistently with the reference interpretation and provided a semantic alignment score together with material-specific mitigation guidance. |

As shown in Table 3, all three systems identified the primary defect as lack of fusion porosity for the representative example. The general-purpose vision language models provided broadly reasonable visual observations and mitigation suggestions, including increasing energy input, improving track overlap, decreasing scan speed, increasing laser power, and reducing hatch spacing. These outputs indicate that general-purpose vision-language models can recognize salient visual patterns in LPBF defect micrographs and generate plausible defect-level interpretations. The proposed LPBF defect agent differs primarily in how it structures and constrains the output. In addition to identifying the defect type, the agent reports a semantic alignment score and links the assessment to IN625-specific mitigation guidance derived from the literature. For example, the agent recommends increasing the laser power, decreasing the scan speed, and maintaining the volumetric energy density within a process window reported for IN625. This output illustrates the role of the ontology-guided knowledge base in translating visual interpretation into structured, material-specific mitigation guidance.

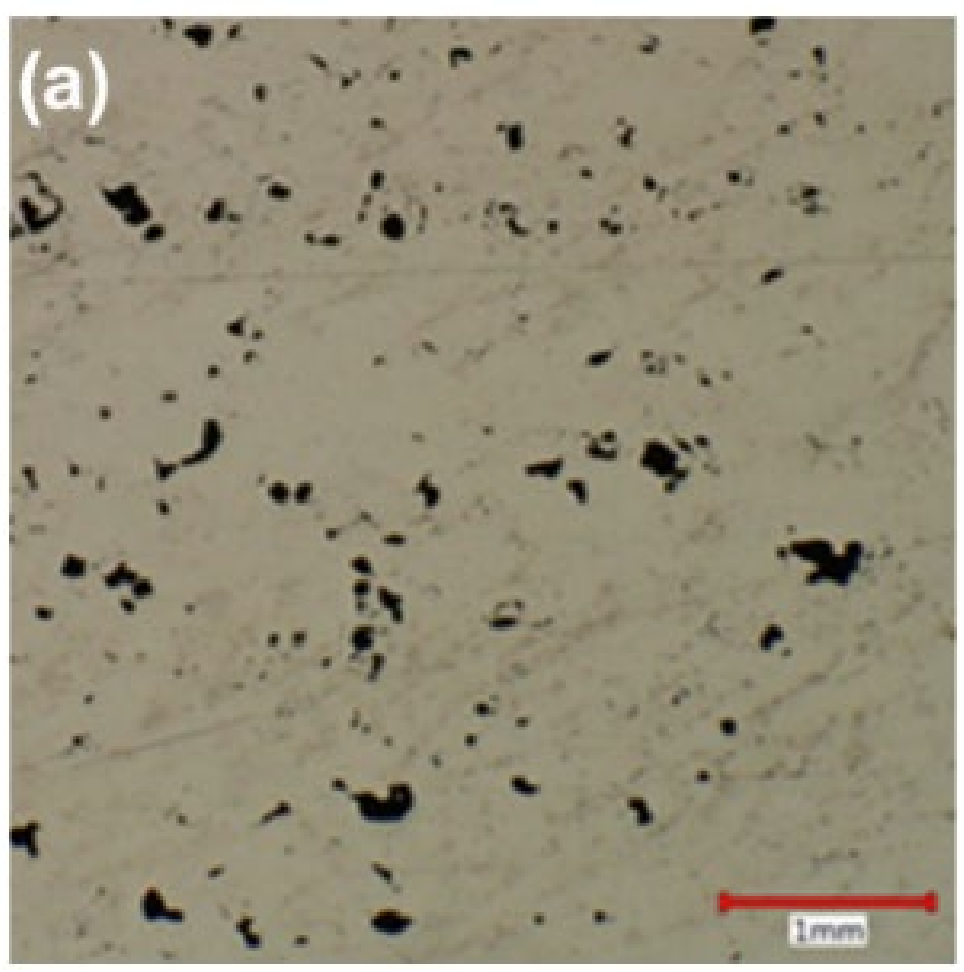

**Figure 13:** Representative LPBF micrograph used as the common visual input for qualitative comparison between the proposed LPBF defect agent and general-purpose vision language models

### 4.2 Ablation study of system components

An ablation study was conducted to evaluate the contributions of the main components of the proposed LPBF defect agent, including the base vision-language model, the dynamic retrieval (RAG) module, and the ontology-guided knowledge base. The evaluation used a strict top-1 macro-averaged evaluation metric to assess the accuracy and usefulness of the proposed framework across defect classes. In multi-class defect classification, global aggregation may conceal weak performance in individual defect classes. Therefore, precision, recall, and F1 score were calculated at the class level and then macro-averaged to provide a class-balanced summary of performance across the evaluated defect morphologies [37].

To assess the diagnostic performance of the proposed knowledge-driven LPBF defect agent, a dataset of high-resolution LPBF microscopy images was compiled from several peer-reviewed studies [27,38–46]. The dataset included representative images of four commonly reported LPBF defect classes: lack of fusion porosity, gas porosity, keyhole porosity, and balling. The final evaluation dataset contained 180 microscopy images from 39 sample parts. Because the images were drawn from multiple published studies, the literature-derived dataset captures defect morphologies reported under varying LPBF process conditions. The labels reported in the source publications were used as reference labels for evaluation, providing a reproducible basis for comparing model outputs across ablation configurations. Using the compiled dataset, four system configurations were evaluated to isolate the contribution of the main components of the proposed agent. Configuration A used the base vision-language model without dynamic retrieval or ontology-guided constraints. Configuration B added dynamic retrieval without the ontology-guided knowledge base. Configuration C used the ontology-guided knowledge base without dynamic retrieval. Configuration D represented the proposed integrated system, combining the ontology-guided knowledge base with targeted retrieval and multimodal reasoning. Table 4 summarizes the resulting accuracy, macro precision, macro recall, and macro F1 scores, and Figure 14 visualizes comparative performance across four system configurations.

**Table 4:** Ablation study results for evaluating the contribution of ontology-guided knowledge representation and dynamic retrieval

| ID | System configuration | Brief description | Accuracy (%) | Precision (%) | Recall (%) | F1 score (%) |
|---|---|---|---|---|---|---|
| A | LLM only | Base vision-language model without dynamic retrieval or ontology-guided knowledge constraints | 64 | 65.2 | 63.8 | 64.5 |
| B | LLM + dynamic RAG | Base model with dynamic retrieval but without ontology-guided knowledge constraints | 12 | 12.5 | 11.8 | 12.1 |
| C | LLM + defect ontology | Base model constrained by the ontology-guided knowledge base without dynamic retrieval | 72 | 74.1 | 71.5 | 72.8 |
| D | Proposed integrated system | Ontology-guided knowledge base with targeted retrieval and multimodal reasoning | 80 | 82.5 | 79.2 | 80.8 |

Configuration A, corresponding to the LLM-only baseline, achieved an overall accuracy of 64.0% and a macro F1 score of 64.5%. This result suggests that the base model could identify some visually salient defect patterns, but its outputs were less consistently constrained by LPBF-specific material and process knowledge.

Configuration B, which added dynamic retrieval without the ontology-guided knowledge base, showed lower performance, with accuracy decreasing to 12.0% and macro F1 score decreasing to 12.1%. This result suggests that unconstrained retrieval can introduce irrelevant or out-of-context information, particularly when retrieved sources concern materials, process parameters, or operating windows that differ from the target IN625 setting. Without an internal knowledge structure to prioritize defect-specific and material-specific information, retrieval may reduce diagnostic consistency rather than improve it.

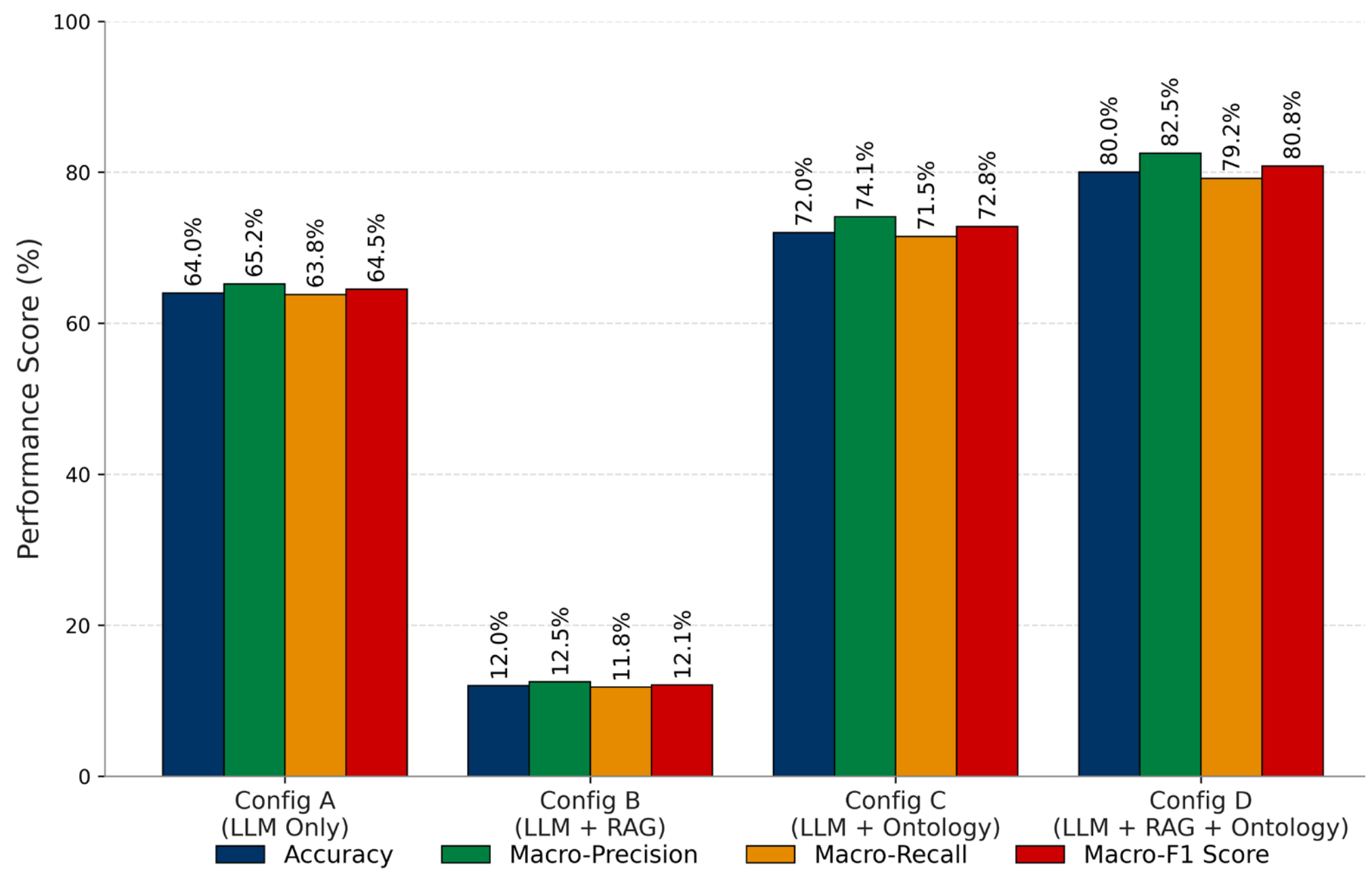

**Figure 14:** Comparison of accuracy, macro precision, macro recall, and macro F1 score across the four evaluated system configurations

Configuration C, which used the ontology-guided knowledge base without dynamic retrieval, improved performance relative to the base model, with accuracy increasing to 72.0% and macro F1 score increasing to 72.8%. This improvement suggests that structured defect knowledge helps constrain the model's reasoning and supports more consistent mapping between observed defect morphology and known LPBF defect classes.

Configuration D, the proposed integrated system, achieved the highest performance among the evaluated configurations, with 80.0% accuracy and a macro F1 score of 80.8%. This result suggests that the combination of ontology-guided defect knowledge, multimodal image interpretation, and targeted retrieval provided the most effective configuration within the current evaluation setting. These findings support the value of structured knowledge representation for improving the consistency of LLM-assisted LPBF defect assessment.

### 4.3 Inter-rater reliability analysis using Cohen's Kappa

An inter-rater reliability analysis was conducted using Cohen's kappa to assess agreement between each system configuration's diagnostic outputs and the literature-derived reference labels. Cohen's kappa ($k$) was selected as the evaluation metric because it accounts for both observed agreement and the agreement expected by chance. The coefficient is calculated as follows:

$$k = \frac{p_0 - p_e}{1 - p_e}$$

Here, $p_0$ represents the observed proportion of agreement between the system output and the reference label and $p_e$ represents the expected proportion of agreement by chance.

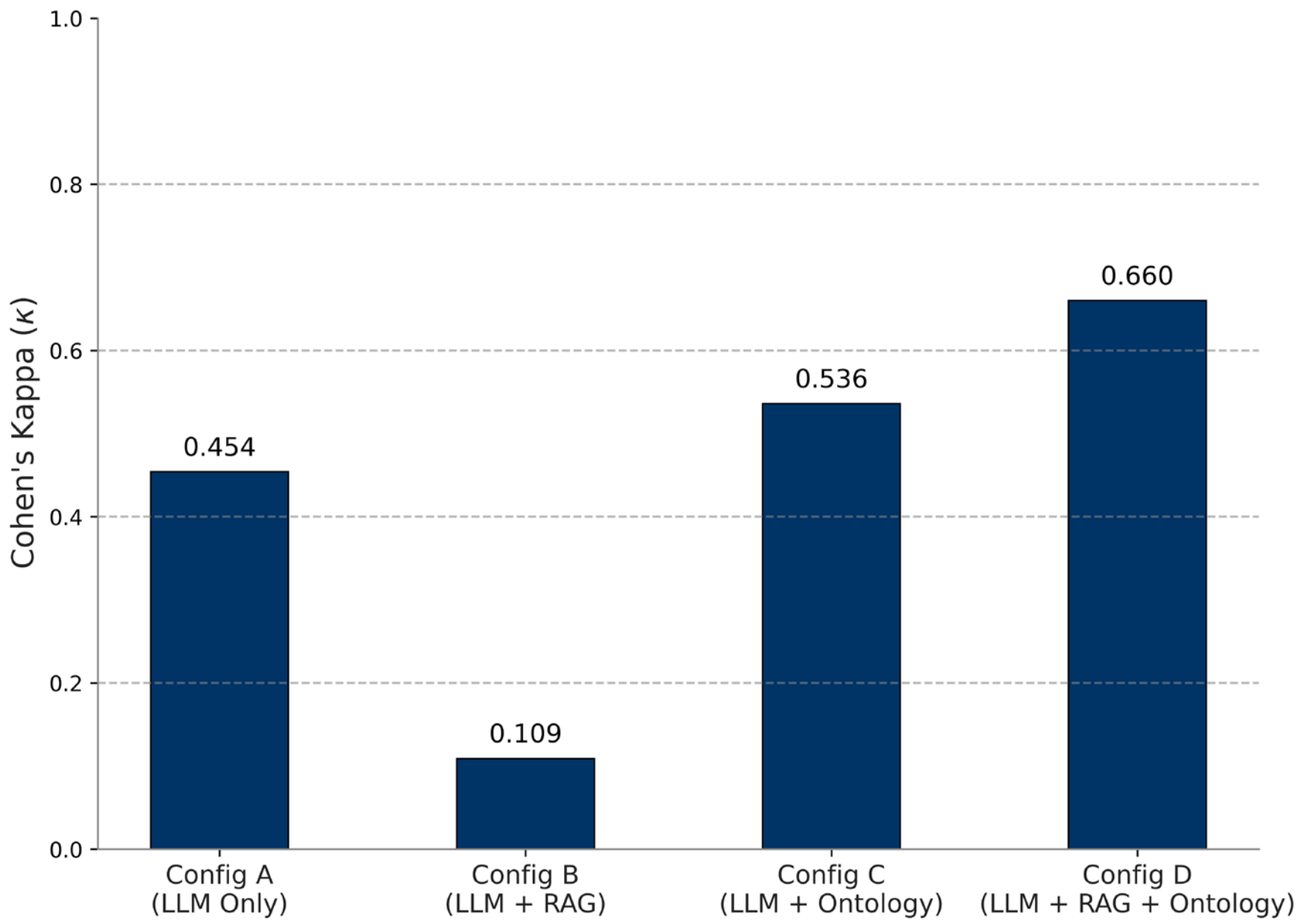

**Figure 15:** Impact of dynamic retrieval and ontology constraints on AI-expert inter-rater reliability

Figure 15 shows the Cohen's kappa values obtained across the four system configurations. The baseline model (Config A) attained a moderate level of agreement with reference labels ($k = 0.454$). The addition of unconstrained dynamic retrieval (Config B) made the diagnostic reliability much worse ($k = 0.109$). It shows that how easily LLMs can be affected by open-source data that is incorrectly provided. However, using the deterministic Application Ontology (Config C) to support the LLM made the reasoning process more stable ($k = 0.536$). Ultimately, the fully integrated design (Config D), which uses the ontology to manage the dynamic retrieval module, had the best performance ($k = 0.66$). Note that any $k > 0.60$ is considered as substantial agreement between two experts [47]. This process empirically illustrates that the suggested ontology-driven constraints are crucial for enhancing the model's diagnostic reasoning to a degree of significant concordance with human domain experts by chance.

## 5. Conclusion

Reliable deployment of metal AM processes such as laser powder bed fusion is constrained by the prevalence of process-induced defects and the knowledge-intensive nature of defect diagnosis and mitigation. Addressing these challenges requires decision-support mechanisms that extend beyond simple information aggregation. Such mechanisms must systematically organize domain knowledge, enable transparent and explainable reasoning, and provide context-specific guidance to support informed engineering decision making. In response, this paper presents an ontology-guided large language model-based defect agent for LPBF defect exploration, interpretation, and mitigation guidance. The proposed framework integrates a hierarchical knowledge base containing 27 LPBF defect types with natural language interaction, fuzzy query handling, supplementary literature retrieval, and multimodal image assessment within a unified workflow. In doing so, it enables users to move beyond static defect lookup to more interactive, explainable defect exploration.

The key contribution of this work lies in connecting structured LPBF defect knowledge with LLM-based interaction and image-assisted reasoning. The developed agent enables users to navigate hierarchical defect categories, retrieve causal information, access mitigation guidance when supporting literature is available, and interpret representative defect micrographs through descriptor-based semantic alignment. Unlike conventional defect classification approaches, the proposed framework emphasizes interpretability by connecting candidate defect assessments to observable image features and structured defect descriptions. The system also supports fuzzy and ambiguous query handling, broadening accessibility for users who may not use exact technical terminology, and incorporates literature-supported retrieval to provide supplementary scholarly context when requested.

The empirical evaluation provided initial evidence of the value of structured knowledge integration. The qualitative comparison with general-purpose vision-language models showed that, for a representative LPBF defect micrograph, the proposed agent produced more structured, material-specific outputs by linking defect interpretation to semantic alignment scores and mitigation guidance. The ablation study further showed that less-constrained system configurations, particularly the retrieval-augmented model without ontology-guided knowledge constraints, were more prone to inconsistent defect assessment in the literature-derived evaluation dataset. By contrast, performance significantly improved when the generative model was supported by deterministic ontology-guided, material-specific knowledge representation. The proposed integrated system achieved the highest performance among the evaluated configurations, with 80.0% diagnostic accuracy and an 80.8% macro F1 score. These findings suggest that ontology-guided knowledge representation can improve the consistency of LLM-assisted LPBF defect assessment, particularly when combined with targeted knowledge retrieval and vision-based defect analysis using foundation models.

This work also defines several important directions for continued development. The current implementation focuses primarily on IN625 because the material-specific mitigation mappings used in the prototype were curated for this alloy. However, the framework can be extended to other LPBF material systems by incorporating material-specific defect mechanisms, process windows, and mitigation guidance into the knowledge base. Similarly, although the present empirical validation is specific to LPBF, the overall framework is modular and could be adapted to other manufacturing processes by replacing the LPBF-specific knowledge base with domain-specific knowledge structures and formally evaluating the resulting system in those settings. Future work will therefore focus on expanding the defect taxonomy, incorporating additional material systems, strengthening causal relationship modeling, and validating the framework using larger, literature-derived and expert-labeled datasets, as well as user studies involving engineers and manufacturing practitioners. Another important direction is integrating in situ monitoring data to extend the framework toward real-time defect interpretation and more predictive decision support.

The proposed LPBF defect agent establishes a systematic and extensible foundation for LPBF defect exploration, interpretable defect assessment, and decision support for defect mitigation in metal additive manufacturing. By integrating curated defect knowledge, an LLM-based interactive platform, supplementary knowledge retrieval, and defect analysis within a unified framework, the proposed LPBF defect agent contributes to the development of more explainable, human-centered decision-support tools and provides a practical step toward more informed defect management in advanced manufacturing environments.

## Declaration of Competing Interest

The authors declare that they have no known competing financial interests or personal relationships that could have influenced the work reported in this paper.